\documentclass[letterpaper]{article} % DO NOT CHANGE THIS
\usepackage{aaai2027}  % DO NOT CHANGE THIS
\usepackage[hyphens]{url}  % DO NOT CHANGE THIS
\usepackage{graphicx} % DO NOT CHANGE THIS
\usepackage{natbib}  % DO NOT CHANGE THIS AND DO NOT ADD ANY OPTIONS TO IT
\usepackage{caption} % DO NOT CHANGE THIS AND DO NOT ADD ANY OPTIONS TO IT
\usepackage{algorithm}
\usepackage{algorithmic}
\newcommand{\eg}{\textit{e.g.}}

\usepackage{newfloat}
\usepackage{listings}
\DeclareCaptionStyle{ruled}{labelfont=normalfont,labelsep=colon,strut=off} % DO NOT CHANGE THIS
\floatstyle{ruled}
\newfloat{listing}{tb}{lst}{}
\floatname{listing}{Listing}

\usepackage{booktabs}
\usepackage[table]{xcolor}
\usepackage{amsmath}
\usepackage{amssymb}
\usepackage{enumitem}
\usepackage{array}
\usepackage{makecell}
\usepackage{multirow}
\providecolor{mygray}{gray}{.92}
\providecolor{ForestGreen}{RGB}{34,139,34}
\providecolor{Forestred}{RGB}{220,50,50}
\providecommand{\multirowsetup}{}
\renewcommand{\multirowsetup}{\centering}

\providecommand{\best}[1]{\textbf{#1}}
\providecommand{\ub}[1]{\textcolor{gray}{#1}}
\providecommand{\src}[1]{\ \texttt{\scriptsize{(#1)}}}

\title{RoRA: Role-Oriented Regional Allocation for Visual Token Pruning in MLLMs
}
\author{
    Qiyanhui Lu\textsuperscript{\rm 1},
    Han Wu\textsuperscript{\rm 2},
    Rongjian Xu\textsuperscript{\rm 1},
    Tingzhang Luo\textsuperscript{\rm 1},
    Cheng Fan\textsuperscript{\rm 1},
    Xinghao Chen\textsuperscript{\rm 3},\\
    Minjing Dong\textsuperscript{\rm 1},
    Jufeng Yang\textsuperscript{\rm 4},
    Jianyuan Guo\textsuperscript{\rm 1}\corresponding
}
\affiliations{
    \textsuperscript{\rm 1}City University of Hong Kong, 
    \textsuperscript{\rm 2}Peking University,
    \textsuperscript{\rm 3}Huawei Technologies,
    \textsuperscript{\rm 4}Nankai University\\
    qiyanhulu2-c@my.cityu.edu.hk, \quad jianyguo@cityu.edu.hk
}

\begin{document}

\maketitle

\begin{abstract}
% Multimodal large language models (MLLMs) encode images as long visual token sequences, making prefilling and KV-cache storage costly. Existing training-free pruning methods select tokens by importance, diversity, or spatial coverage, but treat retained tokens as interchangeable and do not explicitly model which object-related regions are already covered. We introduce RoRA, a training-free framework that reformulates visual token pruning as role-oriented regional evidence allocation. RoRA divides a fixed budget into a protected semantic core, complementary context, and fine-grained detail. It first calibrates text-conditioned attention with a positional prior and a prompt-calibrated object prior, then constructs Attention-Anchored Regions (AARs) from high-confidence anchors as lightweight proxies for already-covered object support. RoRA explores context mainly outside AARs and restores details with a small AAR-guided budget, while confining pairwise similarity to context-stage redundancy filtering. Across LLaVA-1.5, LLaVA-NeXT, Qwen2.5-VL, and Qwen3-VL, RoRA achieves the best normalized average performance among compared training-free methods under matched budgets. At $88.9\%$ pruning, it retains $96.5\%$ of unpruned performance on LLaVA-1.5; on Qwen3-VL, it surpasses D$^2$Pruner by $5.4$ and $5.3$ points at $75\%$ and $90\%$ pruning. A speed-aligned RoRA variant incurs $0.7\,\mathrm{ms}$ selector overhead and provides a $1.33\times$ end-to-end speedup on an NVIDIA H800.
Multimodal large language models (MLLMs) encode images as long visual token sequences, making prefilling and KV-cache storage expensive. Existing training-free pruning methods select tokens by importance, diversity, or spatial coverage, but treat retained tokens as interchangeable and do not explicitly track which object-related regions are already covered. We present RoRA, a training-free framework that casts visual token pruning as role-oriented regional evidence allocation. Given a fixed budget, RoRA partitions tokens into a protected semantic core, complementary context, and fine-grained detail. It first calibrates text-conditioned attention with a positional prior and a prompt-calibrated object prior, then builds Attention-Anchored Regions (AARs) from high-confidence anchors as lightweight proxies for covered object support. Context is explored mainly outside AARs, while a small AAR-guided budget restores local detail; pairwise similarity is used only for context-stage redundancy filtering. Under matched budgets, RoRA consistently outperforms strong training-free baselines across LLaVA and Qwen-VL families, retaining most of the unpruned accuracy even at aggressive pruning ratios, \eg, 96.5\% of full performance at 88.9\% pruning on LLaVA-1.5, and improving over D$^2$Pruner by about 5\% on Qwen3-VL at 75–90\% pruning. At a 66.7\% pruning ratio, RoRA requires only 0.7ms for token selection and reduces end-to-end inference time by 24.6\%, corresponding to a 1.33$\times$ speedup over unpruned inference on an NVIDIA H800.
\noindent\textbf{Code:} \url{https://github.com/LukieLuu/RoRA}
\end{abstract}

\section{Introduction}
\label{sec:introduction}
Multimodal large language models (MLLMs) represent images as long sequences of visual tokens.
High-resolution images, multi-image inputs, and videos can easily yield hundreds to thousands of tokens, accounting for most of the prefilling cost, KV-cache size, and latency in subsequent LLM layers.
Training-free visual token pruning therefore aims to retain only a small subset of visual tokens while preserving the information essential for downstream tasks under a fixed retention ratio, \eg, keeping $r\%$ of the visual sequence.

\begin{figure}[t]
    \centering
    \includegraphics[width=0.48\textwidth]{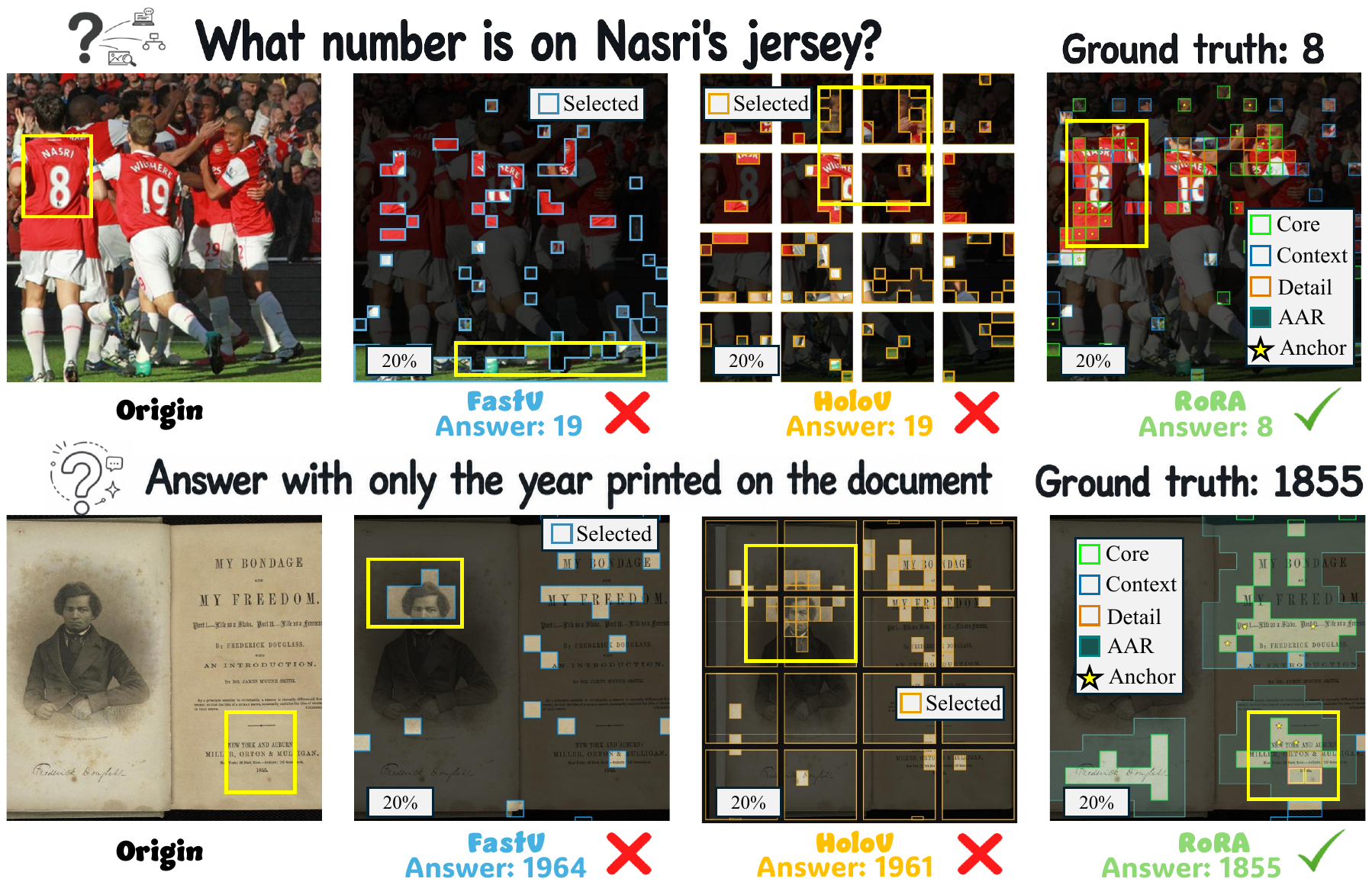}
    \caption{\small{Qualitative comparison of FastV, HoloV, and RoRA under the same visual token budget. FastV is affected by positional attention sinks, while HoloV’s crop-wise allocation assigns tokens to high-variance but task-irrelevant regions. In contrast, RoRA uses AARs to allocate tokens across semantic core, complementary context, and fine-grained detail, thereby preserving the evidence needed to identify the jersey number and document year correctly.}}
    \label{fig:motivation}
    \vspace{-5pt}
\end{figure}

One natural signal for token selection is the text-conditioned attention produced by LLM layers.
However, raw text-conditioned attention is often positionally biased and concentrates around image borders or other systematic hotspots~\cite{holov}. Selecting tokens solely based on attention, as in FastV~\cite{fastv}, therefore leads the retained set to cluster around biased regions while overlooking the true semantic evidence (Figure~\ref{fig:motivation}). As a result, subsequent training-free methods are largely motivated by the same objective: retain tokens that are informative for downstream understanding but would otherwise be discarded by attention-only selection.

Existing remedies fall into two main branches. The first branch does \emph{not} explicitly correct the attention bias.
Instead, it performs \emph{image-level} allocation that promotes non-redundancy and spatial coverage.
DART~\cite{dart} selects diversified tokens relative to pivots, while HoloV~\cite{holov} partitions the image into crops and assigns per-crop quotas using attention and feature variance.
These strategies improve coverage over naive top-$k$ attention.
However, these strategies remain agnostic to the semantic role of each retained token.
In particular, HoloV can over-allocate budget to cluttered background crops: cluttered regions often exhibit high local variance and thus high crop scores, causing many retained tokens to be allocated to spectators or texture-rich edges rather than the queried object (Figure~\ref{fig:motivation}).
The second branch explicitly corrects positional bias before ranking tokens by the debiased attention. D$^2$Pruner~\cite{d2pruner} debiases attention and then suppresses redundancy with a dense token--token similarity graph (\eg, $576\times576$).
Debiasing produces a more reliable attention ranking, but global pairwise modeling is expensive and largely structure-agnostic: without explicitly modeling which regions are already represented by object evidence, the method must compare almost everything to everything to approximate coverage.

Across both branches, a common limitation remains: tokens are selected by importance, diversity, or spatial coverage, but not by the \emph{role} they should play under a fixed token budget.
We argue that answering a multimodal query typically requires three distinct yet complementary types of visual evidence.
First, a small set of high-confidence tokens must lock onto the queried object (\emph{core}); without them, the model loses the primary referent.
Second, once the core object is secured, additional tokens are more useful as complementary scene or secondary-object cues outside the support (\emph{context}), rather than as near-duplicates around the same object.
Third, a few fine-grained tokens are still needed for text, boundaries, and small parts (\emph{detail}), which are easy to miss under coarse importance ranking or crop-level quotas.
Because these roles complement rather than replace one another, allocating the budget with one homogeneous criterion, or with crop quotas that ignore object coverage, cannot deliberately preserve all three.

We propose \textbf{RoRA}, a training-free framework for role-oriented regional evidence allocation. After obtaining debiased attention, we calibrate sample-specific relevance with a lightweight role-aware prompt and protect a semantic \emph{core}.
We then leverage the remaining positional structure as a useful prior: high-confidence anchors and their local neighborhoods on the 2D token grid form Attention-Anchored Regions (AARs).
An AAR is not an exact object mask; it is a lightweight spatial proxy of object-related support already covered by the core.
Conditioned on AARs, RoRA allocates the residual budget asymmetrically: explore complementary \emph{context} mainly outside AARs, and preserve \emph{detail} mainly inside AARs.
Because coverage is made explicit by AARs, redundancy control reduces to comparing candidate context tokens against the protected core, instead of building a full pairwise redundancy matrix over the whole image. This yields both clearer evidence allocation and substantially lower selection overhead, as illustrated in Figure~\ref{fig:motivation}.

We evaluate RoRA on several representative MLLMs including LLaVA-1.5, LLaVA-NeXT, Qwen2.5-VL, and Qwen3-VL. 
RoRA consistently achieves higher normalized average scores than prior
training-free methods under equal retention budgets, while maintaining negligible selection overhead, \eg, 0.7ms per image.
Our contributions are summarized as follows:
\begin{itemize}[nosep, leftmargin=10pt, topsep=0pt, parsep=0pt, partopsep=0pt]
  \item \textbf{Role-aware budget allocation.} We organize retained visual tokens into three complementary roles, and apply different selection criteria under a shared budget.
  \item \textbf{Object region prior.} Beyond debiasing, we exploit a prompt-calibrated signal to identify object-centric regions and stabilize semantic-core protection.
  \item \textbf{Attention-anchored regions for structured pruning.} AARs identify object-related spatial support, enabling outside-AAR context exploration and inside-AAR detail preservation with core-referenced redundancy checks, without requiring dense global similarity graphs.
  \item \textbf{Extensive experiments.} Comparisons and ablations on several MLLMs verify RoRA's accuracy--efficiency gains.

\end{itemize}

\section{Related Work}
\label{sec:related_work}

\textbf{Visual Token Compression in MLLMs.}
Visual token compression has become an important technique for reducing the high inference cost of MLLMs caused by long visual sequences. Existing approaches can be performed before visual features enter the language model or during LLM inference. Vision-side methods reduce tokens through pooling, query-based compression, or token merging before multimodal fusion
~\cite{blip2,deco,tome,llava_prumerge,ipcv}. However, these methods generally lack direct interaction with the textual instruction. In contrast, training-free LLM-side pruning exploits text-conditioned signals during inference to remove visual tokens after they have acquired multimodal
semantics ~\cite{fastv,dart,holov,d2pruner}. We focus on this setting and aim to improve visual token selection under a fixed retention budget.

\noindent\textbf{Training-Free Visual Token Selection.} Existing training-free visual token selection methods mainly rely on three types of signals: \textbf{(i) Attention-based methods.}
FastV and related approaches estimate token importance from text-conditioned attention~\cite{fastv,fastervlm}. However, original text-conditioned attention can contain positional bias, causing important tokens to be confused with model-specific attention patterns. D$^2$Pruner addresses this issue by introducing attention debiasing and combining it with structural diversity modeling ~\cite{d2pruner}.
\textbf{(ii) Redundancy-aware methods.}
Instead of relying only on importance scores, redundancy-aware methods select tokens according to feature similarity or duplication. DART and DivPrune reduce repeated visual representations by encouraging diverse token subsets ~\cite{dart,divprune}. However, these approaches mainly focus on redundancy reduction and do not explicitly distinguish the semantic roles of retained tokens.
\textbf{(iii) Region-aware methods.}
HoloV improves spatial coverage by dividing images into regions and allocating tokens according to regional attention and feature variance ~\cite{holov}. Nevertheless, fixed regional allocation does not explicitly model which object-related evidence has already been covered. Different from previous methods, RoRA treats retained tokens as different evidence roles and performs role-aware allocation among semantic core, complementary context, and fine-grained detail.

\section{Methods}
\label{sec:methods}

\begin{figure*}[t]
    \centering
    \includegraphics[width=\textwidth]{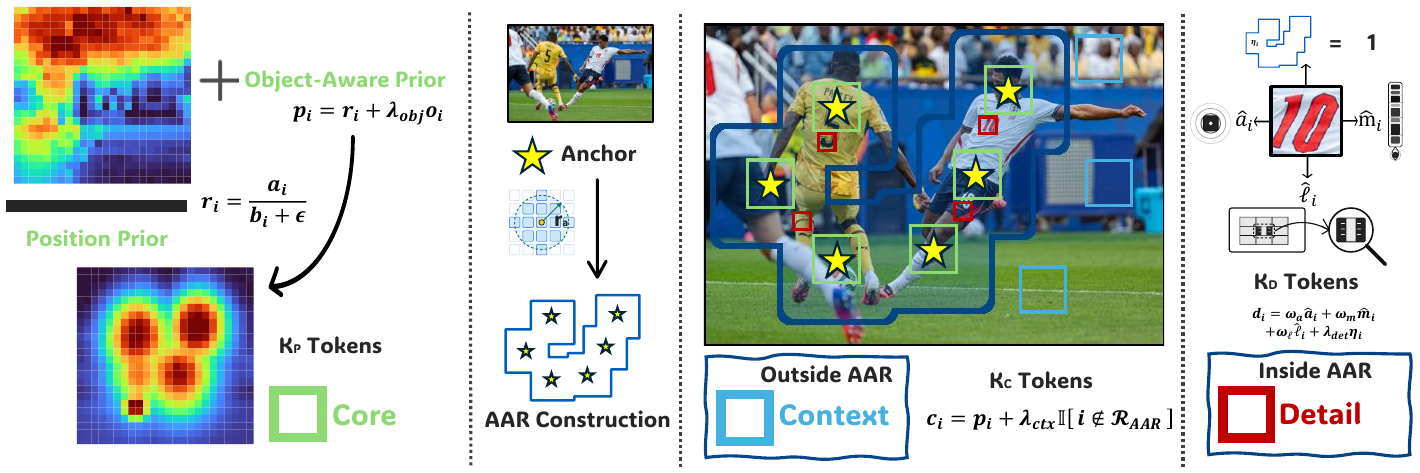}
    \caption{\small{
    \textbf{Overview of RoRA.}
    RoRA decomposes the retained visual token budget into a \textbf{protected
    semantic core}, \textbf{complementary context}, and \textbf{fine-grained details}.
    \textbf{Attention-Anchored Regions} represent object-related spatial support
    already covered by high-confidence evidence, guiding outside-AAR
    context exploration and inside-AAR detail repair.
    }}
    \vspace{-2pt}
    \label{fig:framework}
\end{figure*}

In this section, we present RoRA, a training-free framework for
role-oriented regional evidence allocation.
As illustrated in Figure~\ref{fig:framework}, RoRA first calibrates
text-conditioned visual relevance and protects a high-confidence semantic
core. It then constructs Attention-Anchored Regions (AARs) to represent
object-related spatial support already covered by reliable evidence and
allocates the remaining budget toward complementary context.
Finally, a small-budget detail-repair stage restores fine-grained visual cues.

\subsection{RoRA Framework Overview}
\label{sec:RoRA_overview}
\providecommand{\multirowsetup}{}
\renewcommand{\multirowsetup}{\centering}
\definecolor{mygray}{gray}{.92}
\definecolor{ForestGreen}{RGB}{34,139,34}
\definecolor{Forestred}{RGB}{220,50,50}
\begin{table*}[!ht]
    \centering
    \vspace{-1mm}
    % \hspace{2mm}
    \setlength{\tabcolsep}{3.5pt}
    \renewcommand{\arraystretch}{0.9}
    \footnotesize
    \centering
    \scalebox{0.95}{
    \begin{tabular}{c | c c c c c c c c c | >{\centering\arraybackslash}p{1.0cm}}
        \toprule[1.5pt]
        \textbf{Method} & \textbf{GQA} & \textbf{MMB} & \textbf{MMB-CN} & \textbf{MME} & \textbf{POPE} & \textbf{SQA} & \textbf{VQA}$^{\text{V2}}$ & \textbf{VQA}$^{\text{Text}}$ & \textbf{VizWiz} &  \makecell{\textbf{Avg}.}\\
        \hline
        \rowcolor{mygray}
        LLaVA-1.5-7B & \multicolumn{10}{c}{\textit{Upper Bound, 576 Tokens} \ $\textbf{(100\%)}$}\\
        \textcolor{gray}{Vanilla} & \textcolor{gray}{61.9} & \textcolor{gray}{64.7} & \textcolor{gray}{58.1} & \textcolor{gray}{1862} & \textcolor{gray}{85.9} & \textcolor{gray}{69.5} & \textcolor{gray}{78.4} & \textcolor{gray}{58.2} & \textcolor{gray}{50.0} &  \textcolor{gray}{100.0\%} \\
        \hline

        \rowcolor{mygray}
        LLaVA-1.5-7B & \multicolumn{10}{c}{\textit{Retain 192 Tokens} \ ${(\downarrow 66.7\%)}$}\\
        ToMe \texttt{\scriptsize{(ICLR23)}} & 54.3 & 60.5 & - & 1563 & 72.4 & 65.2 & 68.0 & 52.1 & - & 88.5\% \\
        FastV \texttt{\scriptsize{(ECCV24)}} & 52.7 & 61.2 & 57.0 & 1612 & 64.8 & 67.3 & 67.1 & 52.5 & 50.8 & 90.5\% \\
        MustDrop \texttt{\scriptsize{(arXiv24)}} & 58.2 & 62.3 & 55.8 & 1787 & 82.6 & 69.2 & 76.0 & 56.5 & 51.4 & 97.2\% \\
        LLaVA-PruMerge \texttt{\scriptsize{(ICCV25)}} & 54.3 & 59.6 & 52.9 & 1632 & 71.3 & 67.9 & 70.6 & 54.3 & 50.1 & 91.4\% \\
        PDrop \texttt{\scriptsize{(CVPR25)}} & 57.1 & 63.2 & 56.8 & 1766 & 82.3 & 68.8 & 75.1 & 56.1 & 51.1 & 96.7\% \\
        VisionZip \texttt{\scriptsize{(CVPR25)}} & 59.3 & \textbf{64.5} & \textbf{57.3} & 1767 & \textbf{86.4} & 68.9 & 76.8 & 57.3 & 51.6 & 98.1\% \\
        SparseVLM \texttt{\scriptsize{(ICML25)}} & 57.6 & 62.5 & 53.7 & 1721 & 83.6 & 69.1 & 75.6 & 56.1 & 50.5 & 96.1\% \\
        DART \texttt{\scriptsize{(arxiv25)}}& 60.3 & 64.1 & 56.3 & 1839.8 & 85.2 & \textbf{69.4} & 76.5 & 57.9 & 51.3 & 99.3\% \\
        HoloV \texttt{\scriptsize{(NeurIPS25)}} & 58.7 & 63.5 & 53.6 & 1782.9 & 86.3 & 68.5 & 74.8 & 56.1 & \textbf{51.8} & 97.5\% \\
        D$^2$Pruner \texttt{\scriptsize{(AAAI25)}} & 60.9 & 64.2 & 55.7 & 1854.7 & 85.5 & 69.1 & 77.0 & 58.6 & 51.2 & 99.5\% \\
        RoRA (Ours) & \textbf{61.2} & \textbf{64.5} & 56.2 & \textbf{1861.0} & 85.5 & 69.1 & \textbf{77.1} & \textbf{58.7} & 51.4 & \textbf{99.8\%} \\
        \hline

        \rowcolor{mygray}
        LLaVA-1.5-7B & \multicolumn{10}{c}{\textit{Retain 128 Tokens} \ ${(\downarrow 77.8\%)}$}\\
        ToMe \texttt{\scriptsize{(ICLR23)}} & 52.4 & 53.3 & - & 1343 & 62.8 & 59.6 & 63.0 & 49.1 & -  & 80.4\% \\
        FastV \texttt{\scriptsize{(ECCV24)}} & 49.6 & 56.1 & 56.4 & 1490 & 59.6 & 60.2 & 61.8 & 50.6 & 51.3  & 85.4\%\\
        MustDrop \texttt{\scriptsize{(arXiv24)}} & 56.9 & 61.1 & 55.2 & 1745 & 78.7 & 68.5 & 74.6 & 56.3 & \textbf{52.1} & 95.7\% \\
        LLaVA-PruMerge \texttt{\scriptsize{(ICCV25)}} & 53.3 & 58.1 & 51.7 & 1554 & 67.2 & 67.1 & 68.8 & 54.3 & 50.3 & 89.4\%  \\
        PDrop \texttt{\scriptsize{(CVPR25)}} & 56.0 & 61.1 & 56.6 & 1644 & 82.3 & 68.3 & 72.9 & 55.1 & 51.0  & 94.9\% \\
        VisionZip \texttt{\scriptsize{(CVPR25)}} & 57.6 & 63.4 & \textbf{56.7} & 1768 & 84.7 & 68.8 & 75.6 & 56.8 & 52.0  & 97.2\% \\
        SparseVLM \texttt{\scriptsize{(ICML25)}} & 56.0 & 60.0 & 51.1 & 1696 & 80.5 & 67.1 & 73.8 & 54.9 & 51.4 & 93.8\% \\
        DART \texttt{\scriptsize{(arxiv25)}} & 58.7 & \textbf{63.8} & 54.9 & 1826 & 85.0 & 69.1 & 75.2 & 56.5 & 51.6 & 98.1\% \\
        HoloV \texttt{\scriptsize{(NeurIPS25)}} & 57.3 & 63.2 & 52.8 & 1757.8 & 84.0 & 67.8 & 73.5 & 56.1 & 52.0 & 96.3\% \\
        D$^2$Pruner \texttt{\scriptsize{(AAAI25)}} & \textbf{60.0} & 63.4 & 54.3 & 1846.8 & 85.7 & 69.1 & \textbf{76.3} & 58.3 & 51.5 & 98.8\% \\
        RoRA (Ours) & \textbf{60.0} & 63.7 & 54.9 & \textbf{1849.6} & \textbf{86.1} & \textbf{69.2} & 76.2 & \textbf{58.4} & 51.8 & \textbf{99.1\%} \\
        \hline

        \rowcolor{mygray}
        LLaVA-1.5-7B & \multicolumn{10}{c}{\textit{Retain 64 Tokens} \ ${(\downarrow 88.9\%)}$}\\
        ToMe \texttt{\scriptsize{(ICLR23)}} & 48.6 & 43.7 & - & 1138 & 52.5 & 50.0 & 57.1 & 45.3 & - & 70.1\%  \\
        FastV \texttt{\scriptsize{(ECCV24)}} & 46.1 & 48.0 & 52.7 & 1256 & 48.0 & 51.1 & 55.0 & 47.8 & 50.8  & 76.7\% \\
        MustDrop \texttt{\scriptsize{(arXiv24)}} & 53.1 & 60.0 & 53.1 & 1612 & 68.0 & 63.4 & 69.3 & 54.2 & 51.2  & 90.1\% \\
        LLaVA-PruMerge \texttt{\scriptsize{(ICCV25)}} & 51.9 & 55.3 & 49.1 & 1549 & 65.3 & 68.1 & 67.4 & 54.0 & 50.1  & 87.7\% \\
        PDrop \texttt{\scriptsize{(CVPR25)}} & 41.9 & 33.3 & 50.5 & 1092 & 55.9 & 68.6 & 69.2 & 45.9 & 50.7  & 77.5\% \\
        VisionZip \texttt{\scriptsize{(CVPR25)}} & 55.1 & 60.1 & \textbf{55.4} & 1690 & 77.0 & 69.0 & 72.4 & 55.5 & 52.9  & 94.5\% \\
        SparseVLM \texttt{\scriptsize{(ICML25)}} & 52.7 & 56.2 & 46.1 & 1505 & 75.1 & 62.2 & 68.2 & 51.8 & 50.1 & 87.3\% \\
        DART \texttt{\scriptsize{(arxiv25)}}& 56.1 & 61.6 & 53.4 & 1755 & 82.8 & \textbf{69.1} & 72.0 & 52.7 & 52.0  & 95.2\% \\
        HoloV \texttt{\scriptsize{(NeurIPS25)}} & 55.1 & 60.0 & 51.3 & 1696.8 & 80.5 & 68.9 & 70.8 & 54.6 & \textbf{53.2}  & 94.1\% \\
        D$^2$Pruner \texttt{\scriptsize{(AAAI25)}} & 56.9 & \textbf{62.4} & 52.4 & 1771.9 & 83.3 & 69.0 & 73.2 & 56.2 & 51.7 & 96.2\% \\
        RoRA (Ours) & \textbf{57.5} & 62.0 & 53.1 & \textbf{1773.2} & \textbf{84.2} & 69.0 & \textbf{73.8} & \textbf{56.3} & 51.6 & \textbf{96.5\%} \\

        \hline
        \hline
        \rowcolor{mygray}
        LLaVA-Next-7B & \multicolumn{10}{c}{\textit{Upper Bound, 2880  Tokens} \ $\textbf{(100\%)}$}\\
         \textcolor{gray}{Vanilla} & \textcolor{gray}{64.2} & \textcolor{gray}{67.4} & \textcolor{gray}{60.6} & \textcolor{gray}{1851} & \textcolor{gray}{86.5} & \textcolor{gray}{70.1} & \textcolor{gray}{81.8} & \textcolor{gray}{64.9} & \textcolor{gray}{57.6} &    \textcolor{gray}{100.0\%} \\
          \hline
       \rowcolor{mygray}
        LLaVA-Next-7B & \multicolumn{10}{c}{\textit{Retain 320 Tokens} \ ${(\downarrow 88.9\%)}$} \\

        FastV \texttt{\scriptsize{(ECCV24)}} & 55.9 & 61.6 & 51.9 & 1661 & 71.7 & 62.8 & 71.9 & 55.7 & 53.1  & 88.0\% \\
        
        LLaVA-PruMerge \texttt{\scriptsize{(ICCV25)}} & 53.6 & 61.3 & 55.3 & 1534 & 60.8 & 66.4 & 69.7 & 50.6 & 54.0  & 85.6\% \\

       PDrop \texttt{\scriptsize{(CVPR25)}} & 56.4 & 63.4 & \textbf{56.2} & 1663 & 77.6 & 67.5 & 73.5 & 54.4 & 54.1  & 90.9\% \\

       MustDrop \texttt{\scriptsize{(arXiv24)}} & 57.3 & 62.8 & 55.1 & 1641 & 82.1 & 68.0 & 73.7 & 59.9 & 54.0  & 92.2\% \\

       FasterVLM \texttt{\scriptsize{(ICCV25)}} & 56.9 & 61.6 & 53.5 & 1701 & 83.6 & 66.5 & 74.0 & 56.5 & 52.6 & 91.1\% \\
       
       SparseVLM \texttt{\scriptsize{(ICML25)}} & 56.1 & 60.6 & 54.5 & 1533 & 82.4 & 66.1 & 71.5 & 58.4 & 52.0  & 89.7\%  \\
 
       DART \texttt{\scriptsize{(arxiv25)}} & 61.1 & \textbf{66.3} & 44.1 & 1699.9 & 87.7 & 68.7 & 77.6 & 58.4 & 54.8 & 95.2\%  \\
       HoloV \texttt{\scriptsize{(NeurIPS25)}} & 59.2 & 65.1 & 47.6 & \textbf{1750.2} & 87.4 & 66.5 & 76.3 & 55.7 & \textbf{55.3}  & 94.9\%  \\
       D$^2$Pruner \texttt{\scriptsize{(AAAI25)}} & 61.5 & 66.1 & 41.7 & 1712.7 & 88.0 & 68.4 & \textbf{78.4} & 60.6 & 54.3  & 95.1\%  \\
       RoRA (Ours)& \textbf{61.6} & 66.0 & 42.3 & 1713.6 & \textbf{88.1} & \textbf{68.9} & \textbf{78.4} & \textbf{60.7} & 54.9  & \textbf{95.5\%}  \\

        \bottomrule[1.5pt]
	\end{tabular}}
    \vspace{-1mm}
	\caption{\small{Performance comparisons on LLaVA-1.5-7B and LLaVA-Next-7B across several understanding benchmarks.}}
    \label{tab:llava_main}
     \vspace{-7pt}
\end{table*}

RoRA is applied at an early LLM layer, where visual tokens have
acquired text-conditioned information while pruning can still reduce the
sequence to be processed by all subsequent layers~\cite{fastv,d2pruner}.
Let $\ell_p$ denote the pruning layer. At this LLM layer, the hidden states of visual tokens can be defined as:
\begin{equation}
\mathcal{H}_{V}
=
\left\{
\mathbf{h}_{i}
\right\}_{i=1}^{N}.
\label{eq:hidden_states}
\end{equation}

RoRA retains a total budget of $K$ for visual tokens. Unlike other methods with one homogeneous selection criterion for the entire
visual sequence, it assigns the retained visual tokens with three distinct and non-interchangeable evidence roles:
\begin{equation}
\begin{gathered}
\mathcal{S}
=
\mathcal{P}
\,\dot{\cup}\,
\mathcal{C}
\,\dot{\cup}\,
\mathcal{D},
\\[-1pt]
|\mathcal{S}|=K,
\qquad
K_p+K_c+K_d=K.
\end{gathered}
\label{eq:token_decomposition}
\end{equation}

In this decomposition, $\mathcal{S}$ denotes the final set of retained visual tokens, while
$\mathcal{P}$ encapsulates protected semantic-core tokens that preserve the primary evidence required for query answering. The complementary-context set $\mathcal{C}$ subsumes secondary objects, relational structures, and background context that fall outside the semantic core. Meanwhile, the detail-token set $\mathcal{D}$ recovers fine-grained spatial evidence, including text, small objects, boundaries, and local textures. The symbol $\dot{\cup}$ denotes a disjoint union, so each retained token belongs to exactly one of the three sets. Here,
$K_p=|\mathcal{P}|$, $K_c=|\mathcal{C}|$, and $K_d=|\mathcal{D}|$ denote their respective token counts, while $K$ is the total retained-token budget. Thus, $|\mathcal{S}|=K$ and $K_p+K_c+K_d=K$.

\subsection{Object-Calibrated Attention and Core Protection}
\label{sec:object_calibrated_attention}

Original text-conditioned attention provides an initial estimate of the correlation
between the current text instruction and visual content. However, such methods are vulnerable to positional bias, which systematically favors visual tokens at specific spatial positions. Following the debiasing principle of D$^2$Pruner~\cite{d2pruner}, we calibrate
the instance-level attention with an offline positional-attention
prior. RoRA further incorporates a soft, prompt-calibrated object prior that smoothly steers attention toward regions more likely to contain object-centric evidence.

In practice, we sample 1,000 images from the GQA~\cite{gqa} training split as an unlabeled calibration subset, using only images without questions, answers, or category annotations. For all calibration images, we use the same prompt, ``Please describe the image.'', to estimate the model's positional bias under generic image-description scenarios and obtain a general positional prior. We then repeat the same procedure with the object-inspection prompt, ``List all visible objects in the image.'', to derive relatively weak object-aware prior scores for tokens at each visual position.

This prior is intentionally designed as a subtle regularization term, which is motivated by the fact that object locations naturally vary across images and an excessively strong prior could compromise generalizability.
Yet, a gentle prior suffices to encourage the model to focus on regions with higher semantic likelihood.
Specifically, let $a_i$ denote the attention assigned to the $i$-th visual token, $b_i$ the positional-bias prior, and $o_i$ the object-aware prior. The calibrated attention score is defined as:
\begin{equation}
\begin{aligned}
r_i
&=
\operatorname{Norm}\!\left(
\frac{a_i}{b_i+\epsilon}
\right),\\
p_i
&=
\operatorname{Norm}\!\left(
r_i+\lambda_{\mathrm{obj}}o_i
\right).
\end{aligned}
\label{eq:core_score}
\end{equation}
Here, $\epsilon$ is a numerical stabilizer, $\operatorname{Norm}(\cdot)$ denotes the token-wise score-normalization operation, and $\lambda_{\mathrm{obj}}$ controls the strength of object-aware calibration. Because the object prior is only a weak positional tendency rather than an image-specific object mask, $\lambda_{\mathrm{obj}}$ is intentionally set to a modest value to ensure that sample-specific attention remains the dominant signal. We then select the $K_p$ tokens with the highest $p_i$ scores to form the protected semantic-core set:
\begin{equation}
\mathcal{P}
=
\operatorname{TopK}_{K_p}\!\bigl(
\{p_i\}_{i=1}^{N}
\bigr).
\label{eq:core_set}
\end{equation}
These tokens constitute the semantic backbone of the compressed visual sequence. They are excluded from the candidate pools of subsequent allocation stages and preserved in the final retained set.

\subsection{Attention-Anchored Regional Allocation}
\label{sec:aar_allocation}

The protected semantic core tokens preserve the most reliable semantic evidence. Consequently, selecting the remaining tokens solely by calibrated attention risks over-allocating the token budget to regions already sufficiently covered.
RoRA therefore employs high-confidence tokens as spatial anchors, explicitly modeling the coverage of core evidence to ensure broader context diversity

We select $M_a$ region anchors according to calibrated attention and expand their local neighborhoods on the two-dimensional visual token grid:
\begin{equation}
\begin{gathered}
\mathcal{A} =
\operatorname{TopK}_{M_a}\!\bigl(\{p_i\}_{i=1}^{N}\bigr),\\[-1pt]
\mathcal{R}_{\mathrm{AAR}} =
\bigcup\nolimits_{a\in\mathcal{A}}
\mathcal{N}_{r_a}(a).
\end{gathered}
\label{eq:aar_construction}
\end{equation}
For each anchor, $\mathcal{N}_{r_a}(a)$ represents the local grid region centered at $a$ with radius $r_a$, and $\mathcal{R}_{\mathrm{AAR}}$ denotes the union of all anchor neighborhoods.

Rather than enforcing a fixed token quota for each region, Attention-Anchored Regions (AAR) dynamically adjusts the scores of complementary-context candidates. Once primary object evidence is covered by $\mathcal{P}$, candidates outside AAR receive a small exploration boost, encouraging the remaining budget to cover secondary objects, spatial relations, and scene context.
Based on AAR, we define the complementary-context score of the $i$-th token as
\begin{equation}
c_i
=
\operatorname{Norm}\!\bigl(
p_i
+
\lambda_{\mathrm{ctx}}
\mathbb{I}[\,i\notin\mathcal{R}_{\mathrm{AAR}}\,]
\bigr).
\label{eq:context_score}
\end{equation}
The coefficient $\lambda_{\mathrm{ctx}}$ controls the exploration strength for candidates outside AAR, and $\mathbb{I}[\cdot]$ denotes the indicator function.
When a token falls outside the AAR, its context score receives a small bonus, preventing the remaining token budget from over-concentrating in salient regions that are already adequately covered.

To further reduce redundancy, we use semantic similarity as a lightweight redundancy filter. If the semantic similarity between a candidate token and one already retained token exceeds the threshold $\tau_{\mathrm{sem}}$, this candidate will be skipped, thereby preventing repeated visual evidence from consuming the budget. The resulting complementary-context set is:
\begin{equation}
\mathcal{C}
=
\operatorname{Select}_{\tau_{\mathrm{sem}},K_c}\!\bigl(
\{c_i\mid i\notin\mathcal{P}\}
\bigr).
\label{eq:context_set}
\end{equation}
The operator $\operatorname{Select}_{\tau_{\mathrm{sem}}}$ ranks candidate tokens by descending $c_i$ and skips any candidate whose cosine similarity to the currently retained tokens exceeds $\tau_{\mathrm{sem}}$, until $K_c$ complementary-context tokens have been selected.

\subsection{AAR-Guided Detail Repair}
\label{sec:detail_repair}
\begin{figure*}[t]
    \centering
    \includegraphics[width=\textwidth]{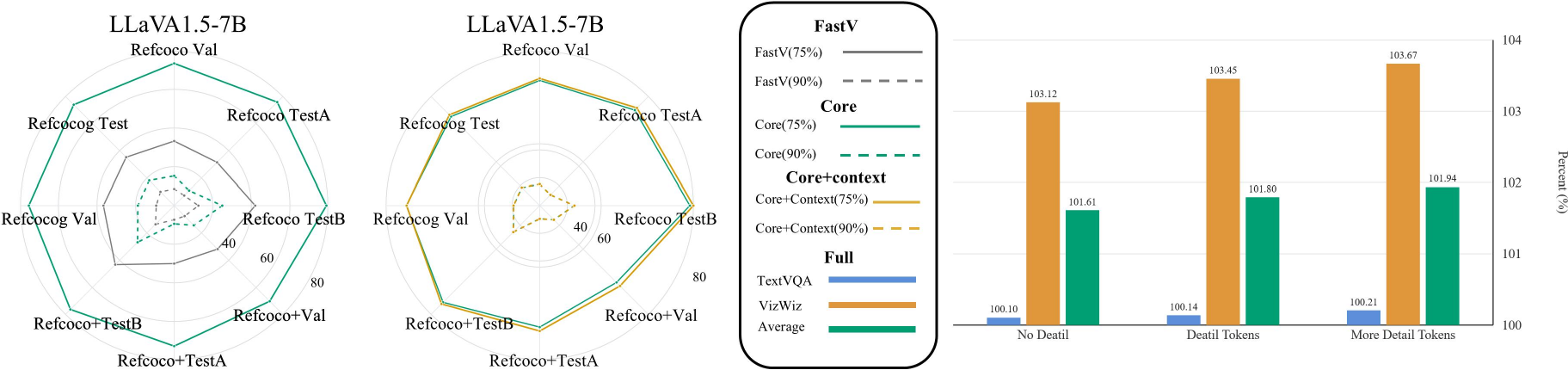}
    \caption{\small{Component-wise ablation on LLaVA-1.5-7B.
    Left: effects of semantic-core protection and
    complementary-context allocation on eight RefCOCO-family localization
    splits at $75\%$ and $90\%$ pruning.
    Right: ablation about detail tokens at $K=128$ on TextVQA and VizWiz.
    }}
    \label{fig:ablation}
\end{figure*}
While the semantic core and complementary context preserve primary objects, relations, and scene information, fine-grained components—such as text, small objects, boundaries, and local textures—occupy few visual patches and often exhibit low global relevance. 
We therefore employ a small-budget detail-repair stage to recover this evidence.

Let $\widehat{a}_i$, $\widehat{m}_i$, and $\widehat{\ell}_i$ denote the attention response, feature magnitude, and local contrast of the $i$-th visual token, respectively:
\begin{equation}
\begin{gathered}
\widehat{a}_i =
\operatorname{Norm}(a_i),
\qquad
\widehat{m}_i =
\operatorname{Norm}\!\left(\|\mathbf{h}_i\|_2\right),\\[-1pt]
\Delta_{ij}=1-\cos(\mathbf{h}_i,\mathbf{h}_j),\\[-1pt]
\widehat{\ell}_i =
\operatorname{Norm}\!\left(
\tfrac{1}{|\mathcal{N}_8(i)|}
\sum\nolimits_{j\in\mathcal{N}_8(i)}
\Delta_{ij}
\right).
\end{gathered}
\label{eq:detail_components}
\end{equation}
For each visual token, $\mathcal{N}_8(i)$ denotes its eight neighborhood on the two-dimensional visual token grid. The attention response represents the attention received by the current visual token; the feature magnitude represents the strength of its hidden-state representation; and the local contrast measures the average feature difference between the current token and its neighboring tokens. Based on these quantities, we define an AAR indicator $\eta_i$ and the detail score of the $i$-th visual token as
\begin{equation}
\begin{gathered}
\eta_i =
\mathbb{I}[\,i\in\mathcal{R}_{\mathrm{AAR}}\,],\\[-1pt]
d_i =
\omega_a\widehat{a}_i
+
\omega_m\widehat{m}_i
+
\omega_{\ell}\widehat{\ell}_i
+
\lambda_{\mathrm{det}}\eta_i.
\end{gathered}
\label{eq:detail_score}
\end{equation}
The coefficients  $\omega_a$, $\omega_m$, and $\omega_{\ell}$ control the contributions of attention response, feature magnitude, and local contrast, respectively, while $\lambda_{\mathrm{det}}$ provides an additional bonus to detail tokens within AAR. Importantly, AAR is not a hard constraint but a soft preference. We select the $K_d$ highest-scoring tokens that have not already been assigned to the semantic-core or complementary-context sets:
\begin{equation}
\mathcal{D}
=
\operatorname{TopK}_{K_d}\!\bigl(
\{d_i\mid i\notin\mathcal{P}\cup\mathcal{C}\}
\bigr).
\label{eq:detail_set}
\end{equation}

Finally, the protected semantic core $\mathcal{P}$, complementary context $\mathcal{C}$, and detail tokens $\mathcal{D}$ jointly form the retained visual token set $\mathcal{S}$.

\section{Experiments}
\label{sec:experiments}

\subsection{Experimental Settings}
\label{sec:experimental_settings}

\definecolor{qwengray}{RGB}{235,235,235}
\begin{table*}[htbp]
    \centering

    \resizebox{0.96\textwidth}{!}{\setlength{\tabcolsep}{1.5pt}
    \renewcommand{\arraystretch}{1.25}
    \begin{tabular}{p{1.8 cm} | ccccccc|c || ccccccc|c}
        \toprule[1.5pt]
        
        % --- Multi-level Header ---
        \smash{\raisebox{-0.85\normalbaselineskip}{\textbf{Method}}} & \multicolumn{8}{c||}{\textbf{Qwen2.5-VL-7B}} & \multicolumn{8}{c}{\textbf{Qwen3-VL-8B}} \\
        \cline{2-17}
        ~ & \textbf{AI2D} & \textbf{MME} & \textbf{T-VQA} & \textbf{NB} & \textbf{RW} & \textbf{TC} & \textbf{SQA} & \textbf{Avg.} & \textbf{BLK} & \textbf{Ill} & \textbf{VR} & \textbf{POPE} & \textbf{T-VQA} & \textbf{CV} & \textbf{MUIR} & \textbf{Avg.} \\
        \hline

        % --- Upper Bound Block ---
        \rowcolor{qwengray}
        ~ & \multicolumn{8}{c||}{\textit{Upper Bound, All Tokens} \ $\textbf{(100\%)}$} & \multicolumn{8}{c}{\textit{Upper Bound, All Tokens} \ $\textbf{(100\%)}$} \\
        \textcolor{gray}{Vanilla}
        & \textcolor{gray}{74.7} & \textcolor{gray}{2246.6} & \textcolor{gray}{76.4} & \textcolor{gray}{24.0} & \textcolor{gray}{27.1} & \textcolor{gray}{53.5} & \textcolor{gray}{73.0} & \textcolor{gray}{100.0\%}
        & \textcolor{gray}{69.1} & \textcolor{gray}{53.7} & \textcolor{gray}{25.6} & \textcolor{gray}{89.1} & \textcolor{gray}{80.9} & \textcolor{gray}{85.9} & \textcolor{gray}{50.9} & \textcolor{gray}{100.0\%} \\
        \hline

        % --- Token Reduction 75% Block ---
        \rowcolor{qwengray}
        ~ & \multicolumn{8}{c||}{\textit{Retain 25\% Tokens in Average} \ $(\downarrow 75\%)$} & \multicolumn{8}{c}{\textit{Retain 25\% Tokens in Average} \ $(\downarrow 75\%)$} \\
        HoloV
        & 71.0 & 1915.2 & 64.7 & 21.8 & 21.7 & 46.2 & 74.8 & 89.2\%
        & 58.3 & 51.0 & 26.5 & 85.0 & 57.1 & 81.0 & 48.4 & 91.1\% \\
        FastV
        & 71.3 & 1915.3 & 73.3 & 23.2 & 30.6 & \textbf{65.6} & 74.1 & 101.4\%
        & 58.3 & \textbf{53.1} & 24.0 & 85.0 & 69.5 & 81.8 & 48.4 & 92.6\% \\
        DART
        & 69.3 & 1900.7 & 66.9 & 25.4 & 27.8 & 54.4 & 73.2 & 96.4\%
        & 55.0 & 51.9 & 25.9 & 85.5 & 69.0 & 81.6 & 49.7 & 93.0\% \\
        DivPrune
        & 21.5 & 2014.4 & 69.8 & 25.1 & 26.1 & 58.4 & 73.4 & 88.6\%
        & 59.1 & 49.2 & 24.5 & \textbf{87.8} & \textbf{72.7} & 81.3 & \textbf{50.5} & 93.6\% \\
        D$^2$Pruner
        & 73.3 & 2080.3 & 73.6 & 28.4 & 31.7 & 60.9 & 75.9 & 105.6\%
        & 58.3 & 49.2 & 24.6 & 85.8 & 70.4 & 82.0 & 49.7 & 92.6\% \\
        RoRA
        & \textbf{73.9} & \textbf{2106.0} & \textbf{73.8} & \textbf{29.4} & \textbf{32.4} & 61.3 & \textbf{76.0} & \textbf{106.9\%}
        & \textbf{62.5} & \textbf{53.1} & \textbf{29.5} & 86.6 & 72.4 & \textbf{82.8} & 50.4 & \textbf{98.0\%} \\
        \hline

        % --- Token Reduction 90% Block ---
        \rowcolor{qwengray}
        ~ & \multicolumn{8}{c||}{\textit{Retain 10\% Tokens in Average} \ $(\downarrow 90\%)$} & \multicolumn{8}{c}{\textit{Retain 10\% Tokens in Average} \ $(\downarrow 90\%)$} \\
        HoloV
        & 64.6 & 1760.3 & 38.7 & 14.0 & 17.3 & 37.5 & \textbf{75.2} & 72.9\%
        & 45.8 & 49.4 & 25.8 & 75.0 & 40.7 & 73.4 & 45.0 & 81.0\% \\
        FastV
        & 65.8 & 1288.9 & 65.0 & 11.6 & 23.5 & \textbf{66.1} & 70.7 & 83.6\%
        & 45.8 & 48.2 & 22.9 & 75.0 & 46.3 & 74.0 & 45.0 & 80.2\% \\
        DART
        & 64.5 & 1610.3 & 56.1 & 17.5 & 24.3 & 44.7 & 72.3 & 82.4\%
        & 45.0 & 49.8 & 23.8 & 77.8 & 49.0 & 72.4 & 47.9 & 82.4\% \\
        DivPrune
        & 19.9 & 1592.4 & 56.9 & 17.7 & 19.3 & 44.3 & 69.2 & 70.6\%
        & 48.3 & 46.6 & 26.0 & \textbf{83.3} & \textbf{59.2} & 74.4 & \textbf{48.5} & 86.6\% \\
        D$^2$Pruner
        & \textbf{72.5} & 1778.5 & 69.5 & 18.3 & 27.1 & 63.9 & 73.6 & 94.8\%
        & 45.0 & 49.2 & 23.0 & 78.9 & 49.8 & 76.6 & 47.1 & 82.6\% \\
        RoRA
        & \textbf{72.5} & \textbf{1780.6} & \textbf{69.6} & \textbf{20.8} & \textbf{27.7} & 64.3 & 73.9 & \textbf{96.7\%}
        & \textbf{49.2} & \textbf{50.8} & \textbf{28.6} & 79.4 & 54.2 & \textbf{77.1} & 47.0 & \textbf{87.9\%} \\
        
        \bottomrule[1.5pt]
    \end{tabular}}
    \vspace{-1mm}
    \caption{\small{Performance comparisons on Qwen2.5-VL-7B-Instruct and Qwen3-VL-8B-Instruct.}}
    \vspace{-7pt}
    \label{tab:qwen_main}
\end{table*}
\paragraph{Models and baselines.}
To evaluate RoRA across different visual-encoding schemes and visual token sequence lengths, we conduct experiments on four representative MLLM backbones: LLaVA-1.5-7B, LLaVA-NeXT-7B, Qwen2.5-VL-7B-Instruct, and Qwen3-VL-8B-Instruct
~\cite{llava,llava_next,qwen25vl,qwen3vl}.
We compare against a broad set of visual token compression baselines, including ToMe, FastV, MustDrop, LLaVA-PruMerge, PDrop, FiCoCo-V, HiRED, FitPrune, VisionZip, SparseVLM, FasterVLM, DART, DivPrune, HoloV, and D$^2$Pruner
~\cite{tome,fastv,mustdrop,llava_prumerge,pdrop,ficoco,hired,fitprune,
visionzip,sparsevlm,fastervlm,dart,divprune,holov,d2pruner}.

\paragraph{Benchmarks.}
For the LLaVA series, we evaluate on GQA, MMBench, MMBench-CN, MME, POPE, ScienceQA, VQAv2, TextVQA, and VizWiz
~\cite{gqa,mmbench,mme,pope,scienceqa,vqav2,textvqa,vizwiz}.
TextVQA and VizWiz are more sensitive to fine-grained visual evidence, whereas benchmarks such as GQA and MME place greater emphasis on global scene understanding and semantic reasoning
~\cite{textvqa,vizwiz,gqa,mme}.
For Qwen2.5-VL-7B-Instruct, we use AI2D, MME, TextVQA, NaturalBench, MME-RealWorld-CN, TextCaps, and ScienceQA-IMG, covering capabilities such as diagram understanding, image-caption generation, fine-grained perception, and multimodal reasoning
~\cite{ai2d,mme,textvqa,naturalbench,mme_realworld,textcaps,scienceqa}.
For the ablation, we additionally evaluate on the
RefCOCO family~\cite{ReferItGame,Mao}.
For Qwen3-VL-8B-Instruct, we further evaluate on BLINK-counting, IllusionVQA, VisRes, POPE, TextVQA, CV-Bench, and MUIRBench to assess pruning robustness in counting, hallucination-sensitive evaluation, visual resolution understanding, and complex multimodal reasoning
~\cite{blink,illusionvqa,visres,pope,textvqa,cvbench,muirbench}.
Unless otherwise specified, we normalize each benchmark score by the corresponding unpruned result and average the normalized scores across benchmarks.

\paragraph{Implementation details.}
RoRA uses one fixed configuration for each
backbone--pruning-ratio pair. Once selected, this configuration is shared
across all tasks and frozen throughout evaluation, without
benchmark-specific tuning. For all competing baselines, we follow the
configurations provided by their official implementations. Complete
hyperparameter settings are reported in the appendix.

\subsection{Main Results}
\label{sec:main_results}
\providecommand{\multirowsetup}{}
\renewcommand{\multirowsetup}{\centering}
\definecolor{mygray}{gray}{.92}
\makeatletter
\@ifundefined{best}{\newcommand{\best}[1]{\textbf{#1}}}{\renewcommand{\best}[1]{\textbf{#1}}}
\@ifundefined{ub}{\newcommand{\ub}[1]{\textcolor{gray}{#1}}}{\renewcommand{\ub}[1]{\textcolor{gray}{#1}}}
\@ifundefined{src}{\newcommand{\src}[1]{\ \texttt{\fontsize{4.8pt}{5.2pt}\selectfont{(#1)}}}}{\renewcommand{\src}[1]{\ \texttt{\fontsize{4.8pt}{5.2pt}\selectfont{(#1)}}}}
\makeatother

\begin{table}[!ht]
\centering
\setlength{\tabcolsep}{1.2pt}
\renewcommand{\arraystretch}{1.00}
\setlength{\extrarowheight}{0.85pt}
{\fontsize{5.6pt}{6.2pt}\selectfont
\resizebox{0.47\textwidth}{!}{%
\begin{tabular}{l | c c c c || c c c c}
\toprule[1.2pt]
\textbf{Methods} & \textbf{Time} & \textbf{Select} & \textbf{Latency} & \textbf{Acc.} & \textbf{Time} & \textbf{Select} & \textbf{Latency} & \textbf{Acc.} \\
\hline
\ub{Upper Bound, 576 Tokens} & \ub{7:35} & \ub{--} & \ub{0.05s} & \ub{100\%} & \ub{7:35} & \ub{--} & \ub{0.05s} & \ub{100\%} \\
\hline
\rowcolor{mygray} LLaVA-1.5-7B & \multicolumn{4}{c||}{\textit{Retain 192 Tokens} \ \textbf{($\downarrow$ 66.70\%)}} & \multicolumn{4}{c}{\textit{Retain 58 Tokens} \ \textbf{($\downarrow$ 90\%)}} \\
FastV\src{ECCV24} & 6:36 & 0.1ms & 0.04s & 75.4\% & 6:44 & 0.1ms & 0.05s & 66.8\% \\
D\raisebox{0.55ex}{\fontsize{3.6pt}{3.6pt}\selectfont 2}Pruner\src{AAAI25} & 19:58 & 71.7ms & 0.13s & 96.8\% & 19:53 & 68.1ms & 0.13s & 87.7\% \\
Faster\src{ICCV25} & 5:51 & 0.1ms & 0.04s & \best{99.2\%} & 5:48 & 0.1ms & 0.04s & 92.3\% \\
MustDrop\src{arXiv24} & 7:24 & 0.2ms & 0.05s & 95.4\% & 9:29 & 0.2ms & 0.06s & 86.6\% \\
DART\src{arXiv25} & 6:17 & 3.2ms & 0.04s & 97.5\% & 6:09 & 2.0ms & 0.04s & 90.6\% \\
HoloV\src{NeurIPS25} & 5:48 & 1.0ms & 0.04s & 98.9\% & 5:38 & 2.4ms & 0.04s & 92.4\% \\
RoRA (Ours) & \best{5:43} & 0.7ms & 0.04s & 98.4\% & \best{5:22} & 0.6ms & 0.04s & \best{92.5\%} \\
\bottomrule[1.2pt]
\end{tabular}%
}}
\vspace{-1.5mm}
\caption{\small{Real inference comparison on LLaVA-1.5 and POPE at 66.7\% and 90\% pruning ratios, measured on a single NVIDIA H800 GPU with batch size 1. All competing methods are evaluated under the same settings using their official open-source implementations.}}
\label{tab:speed}
\vspace{-7pt}
\end{table}
\paragraph{Results on LLaVA models.}
The upper block of Table~\ref{tab:llava_main} reports the results on \textbf{LLaVA-1.5-7B}. With visual token budgets of $K=192$, $128$, and $64$, RoRA retains $99.8\%$, $99.1\%$, and $96.5\%$ of the unpruned model's normalized average performance, respectively, achieving the best result among all compared methods at every matched budget. Moreover, the relative advantage of RoRA increases as pruning becomes more aggressive, indicating that role-aware regional allocation preserves visual evidence more effectively under severely constrained token budgets.

The lower block of Table~\ref{tab:llava_main} further presents results on\textbf{ LLaVA-NeXT-7B}, whose any-resolution visual encoder produces a dynamic number of visual tokens according to the input resolution. When retaining a fixed budget of $K=320$ visual tokens, RoRA achieves a normalized average performance of $95.5\%$, outperforming all compared baselines.

\paragraph{Results on Qwen models.}
Table~\ref{tab:qwen_main} reports results on the Qwen series.
On Qwen2.5-VL-7B, RoRA achieves normalized averages of $106.9\%$ and
$96.7\%$ at $75\%$ and $90\%$ pruning, respectively, outperforming all
compared baselines. On Qwen3-VL-8B, it reaches $98.0\%$ and $87.9\%$,
surpassing D$^2$Pruner by $5.4$ and $5.3$ points, respectively.
These results demonstrate that RoRA generalizes consistently across newer
dynamic-resolution MLLMs.

\subsection{Ablation Studies}
\label{sec:ablation}

We evaluate the semantic core and complementary context on localization
benchmarks, as these components are designed to preserve the queried object
and its surrounding spatial evidence.
The detail stage is evaluated on TextVQA and VizWiz, which are more sensitive
to OCR cues, small objects, and localized visual details.
Localization results are normalized by our unpruned model under the same
prompting, parsing, and IoU-based evaluation protocol, while the detail results
are normalized by the corresponding vanilla upper bounds.

\paragraph{Semantic core and complementary context.}
Figure~\ref{fig:ablation} reports localization results on eight RefCOCO-family
splits~\cite{ReferItGame,Mao}.
All variants retain the same number of visual tokens.
At $K=144$, the normalized average increases from $35.44\%$ for
FastV~\cite{fastv} to $74.44\%$ after introducing the protected semantic
core.
At the more aggressive $K=58$ setting, the average similarly increases from
$9.57\%$ to $17.36\%$.
Core outperforms FastV on every split at both budgets, confirming that
protecting calibrated object-related evidence is essential for localization.

Adding AAR-guided complementary context further improves the average from
$74.44\%$ to $75.03\%$ at $K=144$, and from $17.36\%$ to $17.47\%$
at $K=58$.
It improves seven of eight splits at $K=144$ and introduces no degradation at
$K=58$.
Although the gain is smaller than that of the semantic core, it shows that
allocating part of the fixed budget to previously uncovered spatial evidence
provides additional localization support.

\paragraph{AAR-guided detail repair.}
We evaluate detail repair at $K=128$ on TextVQA~\cite{textvqa} and VizWiz~\cite{vizwiz}. Without detail tokens, the normalized scores are $100.10\%$ on TextVQA and $103.12\%$ on VizWiz, with an average of $101.61\%$. Adding the standard detail configuration increases the average to $101.80\%$, while allocating $18$ detail tokens further improves it to $101.94\%$. The monotonic gains on both benchmarks indicate that detail repair recovers fine-grained evidence not fully captured by the semantic-core and context stages. Its improvement is modest, which is consistent with its intended role as a small-budget complementary module.

\subsection{Efficiency Analysis}
\label{sec:efficiency}

\providecommand{\multirowsetup}{}
\renewcommand{\multirowsetup}{\centering}
\definecolor{mygray}{gray}{.92}
\makeatletter
\@ifundefined{best}{\newcommand{\best}[1]{\textbf{#1}}}{\renewcommand{\best}[1]{\textbf{#1}}}
\@ifundefined{ub}{\newcommand{\ub}[1]{\textcolor{gray}{#1}}}{\renewcommand{\ub}[1]{\textcolor{gray}{#1}}}
\@ifundefined{src}{\newcommand{\src}[1]{\ \texttt{\fontsize{4.8pt}{5.2pt}\selectfont{(#1)}}}}{\renewcommand{\src}[1]{\ \texttt{\fontsize{4.8pt}{5.2pt}\selectfont{(#1)}}}}
\makeatother

\begin{table}[!ht]
\centering
\setlength{\tabcolsep}{1.1pt}
\renewcommand{\arraystretch}{1.02}
\setlength{\extrarowheight}{0.7pt}
{\fontsize{5.6pt}{6.2pt}\selectfont
\resizebox{0.47\textwidth}{!}{%
\begin{tabular}{l|ccccc}
\toprule[1.2pt]
\raisebox{-0.75ex}{\textbf{Methods}} &
\textbf{Prefill $\downarrow$} &
\textbf{Total $\downarrow$} &
\textbf{FLOPs $\downarrow$} &
\textbf{KV Cache $\downarrow$} &
\textbf{Acc. $\uparrow$} \\
& \textbf{(ms/sample)} & \textbf{(ms/sample)} & \textbf{(T)} & \textbf{(MB)} & \textbf{(\%)} \\
\midrule
\ub{Upper Bound, 2928 Tokens} &
\ub{132.8 (1.00$\times$)} &
\ub{133.2 (1.00$\times$)} &
\ub{16.9 (1.00$\times$)} &
\ub{1512 (1.00$\times$)} &
\ub{88.73} \\
\midrule
\rowcolor{mygray}
\multicolumn{6}{l}{\textbf{LLaVA-Next-7B}} \\
D\raisebox{0.55ex}{\fontsize{3.6pt}{3.6pt}\selectfont 2}Pruner\src{AAAI26} (33.4\%) &
126.3 (1.05$\times$) &
125.9 (1.06$\times$) &
6.0 (2.82$\times$) &
526 (2.87$\times$) &
88.82 \\
RoRA (Ours, 33.4\%) &
\best{108.1 (1.23$\times$)} &
\best{106.8 (1.25$\times$)} &
6.0 (2.82$\times$) &
526 (2.87$\times$) &
\best{88.86} \\
D\raisebox{0.55ex}{\fontsize{3.6pt}{3.6pt}\selectfont 2}Pruner\src{AAAI26} (11.2\%) &
69.0 (1.92$\times$) &
68.9 (1.93$\times$) &
2.8 (6.10$\times$) &
198 (7.63$\times$) &
88.00 \\
RoRA (Ours, 11.2\%) &
\best{66.5 (2.00$\times$)} &
\best{65.7 (2.03$\times$)} &
2.8 (6.10$\times$) &
198 (7.63$\times$) &
\best{88.01} \\
\bottomrule[1.2pt]
\end{tabular}%
}}
\vspace{-7pt}
\caption{\small{Accuracy and speed comparison on LLaVA-Next-7B and POPE under 66.7\% and 90\% pruning ratios. We report POPE accuracy, actual prefilling time, total runtime, theoretical FLOPs, and KV cache. Runtime is measured on a single RTX PRO 6000 GPU with batch size 1 and max new tokens 1.}}
\label{tab:next_efficiency}
\vspace{-12pt}
\end{table}
We evaluate LLaVA-1.5-7B on full POPE (9,000 yes-or-no questions) using one NVIDIA H800 GPU, batch size 1, and max new tokens 1. This prefill-dominated protocol measures visual-token selection and LLM prefilling overhead, with $K=192$ and $K=58$ reported in Table~\ref{tab:speed}. Table~\ref{tab:next_efficiency} further evaluates LLaVA-NeXT-7B on one RTX PRO 6000 GPU; due to its much longer visual sequences, selector overhead accounts for a smaller share of total runtime.

At $K=192$, RoRA completes the full evaluation in 5min43s and achieves an average per-sample latency of $0.038\,\mathrm{s}$, both of which are the lowest measured values among the compared methods. Its selector overhead is only $0.704\,\mathrm{ms}$, while it retains $98.4\%$ of the unpruned model's POPE performance. Compared with the unpruned runtime of 7\,min\,35\,s, this corresponds to a 24.6\% reduction in end-to-end inference time. Under the more aggressive $K=58$ setting, RoRA further reduces the total runtime to 5\,min\,22\,s. It achieves not only the lowest inference latency among all compared methods, but also the best accuracy among the compressed models, retaining 92.5\% of the unpruned performance.

RoRA is efficient because token similarity is used only as a lightweight redundancy filter in complementary-context selection: residual candidates highly similar to retained evidence are skipped, instead of injecting pairwise similarity into the primary importance score. Let $N_t<N$ be the residual candidates after fixing the semantic core, and $C$ the number of HoloV crops. RoRA has complexity $\mathcal{O}(N+N_t^2)$, which is much lighter in practice than D$^2$Pruner's $\mathcal{O}(N^2)$ full-graph construction and comparable to HoloV's crop-wise $\mathcal{O}(N^2/C)$.

\section{Conclusion}

In this work, we presented RoRA, a novel training-free visual token pruning framework that views context reduction as role-aware regional evidence allocation. By organizing visual tokens into a protected semantic core, complementary context, and fine-grained detail evidence, RoRA explicitly improves coverage of retained tokens through lightweight Attention-Anchored Regions (AARs). This design enables asymmetric budget allocation, expanding complementary evidence outside attended regions while recovering local details inside them, without constructing costly dense pairwise redundancy graphs. Extensive evaluations across the multiple representative MLLM architectures, including LLaVA and Qwen-VL families show that RoRA achieves a superior accuracy–efficiency profile, consistently outperforming existing training-free baselines across different models even at aggressive pruning ratios.

\bibliography{aaai2027}

\begin{thebibliography}{41}
\providecommand{\natexlab}[1]{#1}

\bibitem[{Alvar et~al.(2025)Alvar, Singh, Akbari, and Zhang}]{divprune}
Alvar, S.~R.; Singh, G.; Akbari, M.; and Zhang, Y. 2025.
\newblock Divprune: Diversity-based visual token pruning for large multimodal models.
\newblock In \emph{Proceedings of the Computer Vision and Pattern Recognition Conference}, 9392--9401.

\bibitem[{Arif et~al.(2025)Arif, Yoon, Nikolopoulos, Vandierendonck, John, and Ji}]{hired}
Arif, K. H.~I.; Yoon, J.; Nikolopoulos, D.~S.; Vandierendonck, H.; John, D.; and Ji, B. 2025.
\newblock Hired: Attention-guided token dropping for efficient inference of high-resolution vision-language models.
\newblock In \emph{Proceedings of the AAAI Conference on Artificial Intelligence}, volume~39, 1773--1781.

\bibitem[{Bai et~al.(2025{\natexlab{a}})Bai, Cai, Chen, Chen, Chen, Cheng, Deng, Ding, Gao, Ge et~al.}]{qwen3vl}
Bai, S.; Cai, Y.; Chen, R.; Chen, K.; Chen, X.; Cheng, Z.; Deng, L.; Ding, W.; Gao, C.; Ge, C.; et~al. 2025{\natexlab{a}}.
\newblock Qwen3-vl technical report.
\newblock \emph{arXiv preprint arXiv:2511.21631}.

\bibitem[{Bai et~al.(2025{\natexlab{b}})Bai, Chen, Liu, Wang, Ge, Song, Dang, Wang, Wang, Tang, Zhong, Zhu, Yang, Li, Wan, Wang, Ding, Fu, Xu, Ye, Zhang, Xie, Cheng, Zhang, Yang, Xu, and Lin}]{qwen25vl}
Bai, S.; Chen, K.; Liu, X.; Wang, J.; Ge, W.; Song, S.; Dang, K.; Wang, P.; Wang, S.; Tang, J.; Zhong, H.; Zhu, Y.; Yang, M.; Li, Z.; Wan, J.; Wang, P.; Ding, W.; Fu, Z.; Xu, Y.; Ye, J.; Zhang, X.; Xie, T.; Cheng, Z.; Zhang, H.; Yang, Z.; Xu, H.; and Lin, J. 2025{\natexlab{b}}.
\newblock Qwen2.5-VL Technical Report.
\newblock arXiv:2502.13923.

\bibitem[{Bolya et~al.(2022)Bolya, Fu, Dai, Zhang, Feichtenhofer, and Hoffman}]{tome}
Bolya, D.; Fu, C.-Y.; Dai, X.; Zhang, P.; Feichtenhofer, C.; and Hoffman, J. 2022.
\newblock Token merging: Your vit but faster.
\newblock \emph{arXiv preprint arXiv:2210.09461}.

\bibitem[{Chen et~al.(2024)Chen, Zhao, Liu, Bai, Lin, Zhou, and Chang}]{fastv}
Chen, L.; Zhao, H.; Liu, T.; Bai, S.; Lin, J.; Zhou, C.; and Chang, B. 2024.
\newblock An image is worth 1/2 tokens after layer 2: Plug-and-play inference acceleration for large vision-language models.
\newblock In \emph{European Conference on Computer Vision}, 19--35. Springer.

\bibitem[{Chen et~al.(2025)Chen, Wen, Wu, Liu, Chen, Ma, Li, He, and Zhang}]{ipcv}
Chen, Y.; Wen, Z.; Wu, Y.; Liu, X.; Chen, S.; Ma, J.; Li, W.; He, C.; and Zhang, L. 2025.
\newblock IPCV: Information-Preserving Compression for MLLM Visual Encoders.
\newblock \emph{arXiv preprint arXiv:2512.18747}.

\bibitem[{Fu et~al.(2026)Fu, Chen, Shen, Qin, Zhang, Lin, Yang, Zheng, Li, Sun et~al.}]{mme}
Fu, C.; Chen, P.; Shen, Y.; Qin, Y.; Zhang, M.; Lin, X.; Yang, J.; Zheng, X.; Li, K.; Sun, X.; et~al. 2026.
\newblock Mme: A comprehensive evaluation benchmark for multimodal large language models.
\newblock \emph{Advances in Neural Information Processing Systems}, 38.

\bibitem[{Fu et~al.(2024)Fu, Hu, Li, Feng, Wang, Lin, Roth, Smith, Ma, and Krishna}]{blink}
Fu, X.; Hu, Y.; Li, B.; Feng, Y.; Wang, H.; Lin, X.; Roth, D.; Smith, N.~A.; Ma, W.-C.; and Krishna, R. 2024.
\newblock Blink: Multimodal large language models can see but not perceive.
\newblock In \emph{European Conference on Computer Vision}, 148--166. Springer.

\bibitem[{Goyal et~al.(2017)Goyal, Khot, Summers-Stay, Batra, and Parikh}]{vqav2}
Goyal, Y.; Khot, T.; Summers-Stay, D.; Batra, D.; and Parikh, D. 2017.
\newblock Making the v in vqa matter: Elevating the role of image understanding in visual question answering.
\newblock In \emph{Proceedings of the IEEE conference on computer vision and pattern recognition}, 6904--6913.

\bibitem[{Gurari et~al.(2018)Gurari, Li, Stangl, Guo, Lin, Grauman, Luo, and Bigham}]{vizwiz}
Gurari, D.; Li, Q.; Stangl, A.~J.; Guo, A.; Lin, C.; Grauman, K.; Luo, J.; and Bigham, J.~P. 2018.
\newblock Vizwiz grand challenge: Answering visual questions from blind people.
\newblock In \emph{Proceedings of the IEEE conference on computer vision and pattern recognition}, 3608--3617.

\bibitem[{Han et~al.(2026)Han, Liu, Zhang, Ding, Chen, Chen, Wang, Yan, and Huang}]{ficoco}
Han, Y.; Liu, X.; Zhang, Z.; Ding, P.; Chen, J.; Chen, H.; Wang, D.; Yan, Q.; and Huang, S. 2026.
\newblock Filter, correlate, compress: Training-free token reduction for mllm acceleration.
\newblock In \emph{Proceedings of the AAAI Conference on Artificial Intelligence}, volume~40, 4601--4609.

\bibitem[{Hudson and Manning(2019)}]{gqa}
Hudson, D.~A.; and Manning, C.~D. 2019.
\newblock Gqa: A new dataset for real-world visual reasoning and compositional question answering.
\newblock In \emph{Proceedings of the IEEE/CVF conference on computer vision and pattern recognition}, 6700--6709.

\bibitem[{Kazemzadeh et~al.(2014)Kazemzadeh, Ordonez, Matten, and Berg}]{ReferItGame}
Kazemzadeh, S.; Ordonez, V.; Matten, M.; and Berg, T. 2014.
\newblock ReferItGame: Referring to Objects in Photographs of Natural Scenes.
\newblock In \emph{Proceedings of the 2014 Conference on Empirical Methods in Natural Language Processing}, 787--798. Association for Computational Linguistics.

\bibitem[{Kembhavi et~al.(2016)Kembhavi, Salvato, Kolve, Seo, Hajishirzi, and Farhadi}]{ai2d}
Kembhavi, A.; Salvato, M.; Kolve, E.; Seo, M.; Hajishirzi, H.; and Farhadi, A. 2016.
\newblock A diagram is worth a dozen images.
\newblock In \emph{European conference on computer vision}, 235--251. Springer.

\bibitem[{Li et~al.(2024)Li, Lin, Peng, Nyandwi, Jiang, Ma, Khanuja, Krishna, Neubig, and Ramanan}]{naturalbench}
Li, B.; Lin, Z.; Peng, W.; Nyandwi, J. d.~D.; Jiang, D.; Ma, Z.; Khanuja, S.; Krishna, R.; Neubig, G.; and Ramanan, D. 2024.
\newblock Naturalbench: Evaluating vision-language models on natural adversarial samples.
\newblock \emph{Advances in Neural Information Processing Systems}, 37: 17044--17068.

\bibitem[{Li et~al.(2023{\natexlab{a}})Li, Li, Savarese, and Hoi}]{blip2}
Li, J.; Li, D.; Savarese, S.; and Hoi, S. 2023{\natexlab{a}}.
\newblock Blip-2: Bootstrapping language-image pre-training with frozen image encoders and large language models.
\newblock In \emph{International conference on machine learning}, 19730--19742. PMLR.

\bibitem[{Li et~al.(2023{\natexlab{b}})Li, Du, Zhou, Wang, Zhao, and Wen}]{pope}
Li, Y.; Du, Y.; Zhou, K.; Wang, J.; Zhao, X.; and Wen, J.-R. 2023{\natexlab{b}}.
\newblock Evaluating object hallucination in large vision-language models.
\newblock In \emph{Proceedings of the 2023 conference on empirical methods in natural language processing}, 292--305.

\bibitem[{Liu et~al.(2024{\natexlab{a}})Liu, Li, Li, and Lee}]{llava}
Liu, H.; Li, C.; Li, Y.; and Lee, Y.~J. 2024{\natexlab{a}}.
\newblock Improved baselines with visual instruction tuning.
\newblock In \emph{Proceedings of the IEEE/CVF conference on computer vision and pattern recognition}, 26296--26306.

\bibitem[{Liu et~al.(2024{\natexlab{b}})Liu, Li, Li, Li, Zhang, Shen, and Lee}]{llava_next}
Liu, H.; Li, C.; Li, Y.; Li, B.; Zhang, Y.; Shen, S.; and Lee, Y.~J. 2024{\natexlab{b}}.
\newblock Llavanext: Improved reasoning, ocr, and world knowledge.

\bibitem[{Liu et~al.(2024{\natexlab{c}})Liu, Shi, Hong, Hu, Yin, and Zhang}]{mustdrop}
Liu, T.; Shi, L.; Hong, R.; Hu, Y.; Yin, Q.; and Zhang, L. 2024{\natexlab{c}}.
\newblock Multi-stage vision token dropping: Towards efficient multimodal large language model.
\newblock \emph{arXiv preprint arXiv:2411.10803}.

\bibitem[{Liu et~al.(2024{\natexlab{d}})Liu, Duan, Zhang, Li, Zhang, Zhao, Yuan, Wang, He, Liu et~al.}]{mmbench}
Liu, Y.; Duan, H.; Zhang, Y.; Li, B.; Zhang, S.; Zhao, W.; Yuan, Y.; Wang, J.; He, C.; Liu, Z.; et~al. 2024{\natexlab{d}}.
\newblock Mmbench: Is your multi-modal model an all-around player?
\newblock In \emph{European conference on computer vision}, 216--233. Springer.

\bibitem[{Lu et~al.(2022)Lu, Mishra, Xia, Qiu, Chang, Zhu, Tafjord, Clark, and Kalyan}]{scienceqa}
Lu, P.; Mishra, S.; Xia, T.; Qiu, L.; Chang, K.-W.; Zhu, S.-C.; Tafjord, O.; Clark, P.; and Kalyan, A. 2022.
\newblock Learn to explain: Multimodal reasoning via thought chains for science question answering.
\newblock \emph{Advances in neural information processing systems}, 35: 2507--2521.

\bibitem[{Mao et~al.(2016)Mao, Huang, Toshev, Camburu, Yuille, and Murphy}]{Mao}
Mao, J.; Huang, J.; Toshev, A.; Camburu, O.; Yuille, A.~L.; and Murphy, K. 2016.
\newblock Generation and Comprehension of Unambiguous Object Descriptions.
\newblock In \emph{Proceedings of the IEEE Conference on Computer Vision and Pattern Recognition}, 11--20.

\bibitem[{Shahgir et~al.(2024)Shahgir, Sayeed, Bhattacharjee, Ahmad, Dong, and Shahriyar}]{illusionvqa}
Shahgir, H.~S.; Sayeed, K.~S.; Bhattacharjee, A.; Ahmad, W.~U.; Dong, Y.; and Shahriyar, R. 2024.
\newblock Illusionvqa: A challenging optical illusion dataset for vision language models.
\newblock \emph{arXiv preprint arXiv:2403.15952}.

\bibitem[{Shang et~al.(2025)Shang, Cai, Xu, Lee, and Yan}]{llava_prumerge}
Shang, Y.; Cai, M.; Xu, B.; Lee, Y.~J.; and Yan, Y. 2025.
\newblock Llava-prumerge: Adaptive token reduction for efficient large multimodal models.
\newblock In \emph{Proceedings of the IEEE/CVF International Conference on Computer Vision}, 22857--22867.

\bibitem[{Sidorov et~al.(2020)Sidorov, Hu, Rohrbach, and Singh}]{textcaps}
Sidorov, O.; Hu, R.; Rohrbach, M.; and Singh, A. 2020.
\newblock Textcaps: a dataset for image captioning with reading comprehension.
\newblock In \emph{European conference on computer vision}, 742--758. Springer.

\bibitem[{Singh et~al.(2019)Singh, Natarajan, Shah, Jiang, Chen, Batra, Parikh, and Rohrbach}]{textvqa}
Singh, A.; Natarajan, V.; Shah, M.; Jiang, Y.; Chen, X.; Batra, D.; Parikh, D.; and Rohrbach, M. 2019.
\newblock Towards vqa models that can read.
\newblock In \emph{Proceedings of the IEEE/CVF conference on computer vision and pattern recognition}, 8317--8326.

\bibitem[{Tong et~al.(2024)Tong, Brown, Wu, Woo, Middepogu, Akula, Yang, Yang, Iyer, Pan et~al.}]{cvbench}
Tong, S.; Brown, E.; Wu, P.; Woo, S.; Middepogu, M.; Akula, S.~C.; Yang, J.; Yang, S.; Iyer, A.; Pan, X.; et~al. 2024.
\newblock Cambrian-1: A fully open, vision-centric exploration of multimodal llms.
\newblock \emph{Advances in Neural Information Processing Systems}, 37: 87310--87356.

\bibitem[{T{\"o}rtei et~al.(2026)T{\"o}rtei, Dahou, Huynh, Para, Khac, Singh, Chaybouti, and Narayan}]{visres}
T{\"o}rtei, B.~M.; Dahou, Y.; Huynh, N.~D.; Para, W.~R.; Khac, P. H.~L.; Singh, A.; Chaybouti, S.; and Narayan, S. 2026.
\newblock VisRes Bench: On Evaluating the Visual Reasoning Capabilities of VLMs.
\newblock In \emph{Proceedings of the IEEE/CVF Conference on Computer Vision and Pattern Recognition}, 33185--33195.

\bibitem[{Wang et~al.(2025)Wang, Fu, Huang, Li, Liu, Liu, Ma, Xu, Zhou, Zhang et~al.}]{muirbench}
Wang, F.; Fu, X.; Huang, J.~Y.; Li, Z.; Liu, Q.; Liu, X.; Ma, M.~D.; Xu, N.; Zhou, W.; Zhang, K.; et~al. 2025.
\newblock Muirbench: A comprehensive benchmark for robust multi-image understanding.
\newblock In \emph{International Conference on Learning Representations}, volume 2025, 62624--62650.

\bibitem[{Wen et~al.(2025)Wen, Gao, Wang, Zhang, Zhang, Li, He, and Zhang}]{dart}
Wen, Z.; Gao, Y.; Wang, S.; Zhang, J.; Zhang, Q.; Li, W.; He, C.; and Zhang, L. 2025.
\newblock Stop Looking for “Important Tokens” in Multimodal Language Models: Duplication Matters More.
\newblock In \emph{Proceedings of the 2025 Conference on Empirical Methods in Natural Language Processing}, 9972--9991.

\bibitem[{Xing et~al.(2025)Xing, Huang, Dong, Lu, Zhang, Zang, Cao, He, Wang, Wu et~al.}]{pdrop}
Xing, L.; Huang, Q.; Dong, X.; Lu, J.; Zhang, P.; Zang, Y.; Cao, Y.; He, C.; Wang, J.; Wu, F.; et~al. 2025.
\newblock Conical visual concentration for efficient large vision-language models.
\newblock In \emph{Proceedings of the IEEE/CVF Conference on Computer Vision and Pattern Recognition}, 14593--14603.

\bibitem[{Yang et~al.(2025)Yang, Chen, Tian, Wang, Li, Yu, and Jia}]{visionzip}
Yang, S.; Chen, Y.; Tian, Z.; Wang, C.; Li, J.; Yu, B.; and Jia, J. 2025.
\newblock Visionzip: Longer is better but not necessary in vision language models.
\newblock In \emph{Proceedings of the IEEE/CVF Conference on Computer Vision and Pattern Recognition}, 19792--19802.

\bibitem[{Yao et~al.(2024)Yao, Li, Ren, Wang, Liu, Sun, and Hou}]{deco}
Yao, L.; Li, L.; Ren, S.; Wang, L.; Liu, Y.; Sun, X.; and Hou, L. 2024.
\newblock Deco: Decoupling token compression from semantic abstraction in multimodal large language models.
\newblock \emph{arXiv preprint arXiv:2405.20985}.

\bibitem[{Ye et~al.(2025)Ye, Wu, Lin, and Zhou}]{fitprune}
Ye, W.; Wu, Q.; Lin, W.; and Zhou, Y. 2025.
\newblock Fit and prune: Fast and training-free visual token pruning for multi-modal large language models.
\newblock In \emph{Proceedings of the AAAI Conference on Artificial Intelligence}, volume~39, 22128--22136.

\bibitem[{Zhang et~al.(2026)Zhang, Yu, Wu, Wen, Yan, Ding, Qi, and Zhang}]{d2pruner}
Zhang, E.; Yu, F.; Wu, A.; Wen, Z.; Yan, K.; Ding, S.; Qi, B.; and Zhang, L. 2026.
\newblock D$^2$Pruner: Debiased Importance and Structural Diversity for MLLM Token Pruning.
\newblock In \emph{Proceedings of the AAAI Conference on Artificial Intelligence}, volume~40, 12412--12420.

\bibitem[{Zhang et~al.(2025{\natexlab{a}})Zhang, Cheng, Lu, Zhang, Zhuo, Cao, Guo, She, and Zhang}]{fastervlm}
Zhang, Q.; Cheng, A.; Lu, M.; Zhang, R.; Zhuo, Z.; Cao, J.; Guo, S.; She, Q.; and Zhang, S. 2025{\natexlab{a}}.
\newblock Beyond text-visual attention: Exploiting visual cues for effective token pruning in vlms.
\newblock In \emph{Proceedings of the IEEE/CVF International Conference on Computer Vision}, 20857--20867.

\bibitem[{Zhang et~al.(2024)Zhang, Fan, Ma, Zheng, Huang, Cheng, Gudovskiy, Okuno, Nakata, Keutzer et~al.}]{sparsevlm}
Zhang, Y.; Fan, C.-K.; Ma, J.; Zheng, W.; Huang, T.; Cheng, K.; Gudovskiy, D.; Okuno, T.; Nakata, Y.; Keutzer, K.; et~al. 2024.
\newblock Sparsevlm: Visual token sparsification for efficient vision-language model inference.
\newblock \emph{arXiv preprint arXiv:2410.04417}.

\bibitem[{Zhang et~al.(2025{\natexlab{b}})Zhang, Zhang, Tian, Fu, Zhang, Wu, Li, Wang, Wen, Zhang et~al.}]{mme_realworld}
Zhang, Y.; Zhang, H.; Tian, H.; Fu, C.; Zhang, S.; Wu, J.; Li, F.; Wang, K.; Wen, Q.; Zhang, Z.; et~al. 2025{\natexlab{b}}.
\newblock Mme-realworld: Could your multimodal llm challenge high-resolution real-world scenarios that are difficult for humans?
\newblock In \emph{International Conference on Learning Representations}, volume 2025, 89655--89701.

\bibitem[{Zou et~al.(2026)Zou, Lu, Wang, Yan, Lyu, Zheng, Zhang, and Hu}]{holov}
Zou, X.; Lu, D.; Wang, Y.; Yan, Y.; Lyu, Y.; Zheng, X.; Zhang, L.; and Hu, X. 2026.
\newblock Don't Just Chase “Highlighted Tokens” in MLLMs: Revisiting Visual Holistic Context Retention.
\newblock \emph{Advances in Neural Information Processing Systems}, 38: 39800--39832.

\end{thebibliography}

\end{document}